# Enhancing Wind Speed and Wind Power Forecasting Using Shape-Wise Feature Engineering: A Novel Approach for Improved Accuracy and Robustness

[1]Mulomba Mukendi Christian, [2]Yun Seon Kim *, [3]Hyebong Choi, [4]Jaeyoung Lee, [5]SongHee You

[1]*PhD candidate, Dept. of Advanced Convergence, Handong Global Univ., Korea*
*mulombachristian@handong.ac.kr*
[2]*Associate Prof., School of Global Entrepreneurship and Information Communication Technology, Handong Global Univ., Korea*
*sean0831@handong.edu*
[3]*Asssociate Prof., School of Global Entrepreneurship and Information Communication Technology, Handong Global Univ., Korea*
*hbchoi@handong.edu,*
[4]*Prof., School of Mechanical control and Engineering, Handong Global Univ., Korea*
*jylee@handong.edu*
[5]*Assistant Prof., School of Spatial Environment System Engineering, Handong Global Univ., Korea*
*shyou@handong.edu*

*Abstract*

 *Accurate prediction of wind speed and power is vital for enhancing the efficiency of wind energy systems. Numerous solutions have been implemented to date, demonstrating their potential to improve forecasting. Among these, deep learning is perceived as a revolutionary approach in the field. However, despite their effectiveness, the noise present in the collected data remains a significant challenge. This noise has the potential to diminish the performance of these algorithms, leading to inaccurate predictions. In response to this, this study explores a novel feature engineering approach. This approach involves altering the data input shape in both Convolutional Neural Network-Long Short-Term Memory (CNN-LSTM) and Autoregressive models for various forecasting horizons. The results reveal substantial enhancements in model resilience against noise resulting from step increases in data. The approach could achieve an impressive 83% accuracy in predicting unseen data up to the 24th steps. Furthermore, this method consistently provides high accuracy for short, mid, and long-term forecasts, outperforming the performance of individual models. These findings pave the way for further research on noise reduction strategies at different forecasting horizons through shape-wise feature engineering.*

## 1. INTRODUCTION

The increasing global reliance on renewable energy sources has prompted a growing need for accurate forecasting of wind speed and wind power, which are fundamental factors in the operation, design, and success of wind energy systems [1], [2] . However, the volatile and changeable nature of wind energy necessitates robust forecasting methods that consider energy storage, load constraints, and other relevant factors [3]. Effective forecasting has a significant impact on cost optimization and overall energy management systems, leading to the development of numerous solutions aimed at providing accurate and reliable predictions [3].

In recent years, the field of wind speed and wind power forecasting has seen advancements in various areas. One notable trend involves the utilization of machine learning techniques, particularly deep learning, which have the capacity to incorporate diverse data sources such as weather forecasts, historical data, and real-time measurements. By modeling the intricate relationships between meteorological variables and wind speed or power output, these techniques offer the potential for highly accurate and reliable forecasts. However, the computational complexity associated with these models has led to the widespread adoption of hybrid models that combine physical models with machine learning algorithms, enabling accurate forecasting while managing computational resources effectively [3], [4]. Furthermore, advancements in weather modeling and sensing technology, such as satellite imagery, ground-based sensors, and atmospheric modeling techniques, have significantly improved the resolution and accuracy of weather forecasts. These advancements have been rapidly integrated into weather forecasting practices, enhancing the quality of wind speed and power predictions [5]. Real-time data analysis, involving the use of sensors and monitoring equipment to collect data in real-time and applying machine learning algorithms or other techniques for analysis, has gained importance in wind speed and power forecasting. This approach enables highly accurate and timely forecasts by leveraging real-time data streams [6]. Numerous studies have contributed to our understanding and application of various techniques in wind speed and wind power forecasting. For instance, Ding et al. [7] proposed a methodology based on gated recurrent unit neural networks for correcting wind speed errors in numerical weather prediction, which showed improved short-term wind forecasting accuracy. Ensemble learning techniques were explored by Ribeiro et al. [8] for very short-term and short-term wind power forecasting, demonstrating superior performance compared to individual forecasting models. Similarly, an optimal ensemble method for one-day-ahead hourly wind power forecasting was presented in [9], yielding more accurate predictions than other methods examined in the study. Integrated models, combining multiple forecasting approaches, have also shown promising results. [10] highlighted the successful predictive ability of integrated models compared to individual models considered in isolation. Additionally, [11] proposed a hybrid CNN-LSTM model for medium- and long-term power load forecasting, showing improved noise handling capabilities compared to existing approaches.

Numerous studies have investigated the prediction of wind speed and wind power using advanced machine learning algorithms. However, many of these studies have relied on univariate methods, which have limitations in capturing the complex relationships between the target variables and other relevant variables. This highlights the need for bivariate or multivariate methods to improve the accuracy and effectiveness of wind speed and power predictions [12].

This study aims to address this need by conducting a bivariate analysis of wind speed and power, specifically focusing on multiple time steps and simultaneous prediction of both variables using a single model. The study proposes a novel combination of the hybrid Convolutional Neural Network (CNN) and Long Short-Term Memory (LSTM) models, known as CNN-LSTM, along with an autoregressive linear model. The feature engineering process employed in this study involves differencing shaping of the input data, which helps mitigate noise in the dataset and leads to more accurate predictions.

The structure of the paper is as follows: Section 1 provides the introduction, highlighting the research gap and the proposed approach. Section 2 describes the methodology, outlining the use of the CNN-LSTM and autoregressive models and the feature engineering process. Section 3 presents the results and discussion, analyzing the performance of the proposed approach. Finally, Section 4 concludes the paper and suggests future directions for research in this field.

## 2. METHODOLOGY

### 2.1. Theoretical background of bivariate multistep timeseries forecasting

Bivariate multistep time series forecasting is a forecasting technique that involves predicting two variables simultaneously for multiple time steps into the future. This approach finds applications in various disciplines, such as finance, economics, and meteorology. By utilizing the historical values of both variables, the model aims to establish the relationships between them and generate accurate predictions for future time steps. The key objective of bivariate multistep forecasting is to understand and capture the dependencies and interactions between the two variables over time. By considering the historical patterns and dynamics between the variables, the model can make informed predictions for multiple time steps ahead[13]. This technique offers valuable insights for various domains where accurate predictions of multiple variables are critical for effective planning, risk management, and decision-making processes [12].

This method can be described by the equation (1):

$$\left(Y1_{(t+h)}, Y2_{(t+h)}\right) = f\left(Y1_{(t)}, Y2_{(t)}, Y1_{(t-1)}, Y2_{(t-1)}, \dots, Y1_{(t-p+1)}, Y2_{(t-p+1)}\right) \quad [1]$$

where:

$\left(Y1_{(t+h)}, Y2_{(t+h)}\right)$ is the vector of forecasted values of the response variables Y1 and Y2 at time t+h.
$p$ is the number of lagged observations of *Y1* and *Y2* included in the model.
$f()$ is the function that maps the values of *Y1* and *Y2* at the previous p time points to the vector of forecasted values of *Y1* and *Y2* at time *t+h*. This function can be linear or neural network based.

Forecasting bivariate multistep time series involves various methods, such as Autoregressive integrated moving average (ARIMA) models [14], Vector autoregression (VAR) [15], State space models [16], Neural network-based models [21], and Regression-based models [17], among others. In this study, a combination of neural network-based and regression-based models is employed, and their performance is analyzed. The specific details of this approach and its implementation are discussed in the subsequent section.

### 2.2. Data

The dataset used in this study consists of 10-minute interval recordings from Jeju Island in South Korea. It includes the following variables: wind power, wind direction, and wind speed [18]. Figure 1 and Figure 2 present visual representations of the dataset and highlight the target variables to predict.

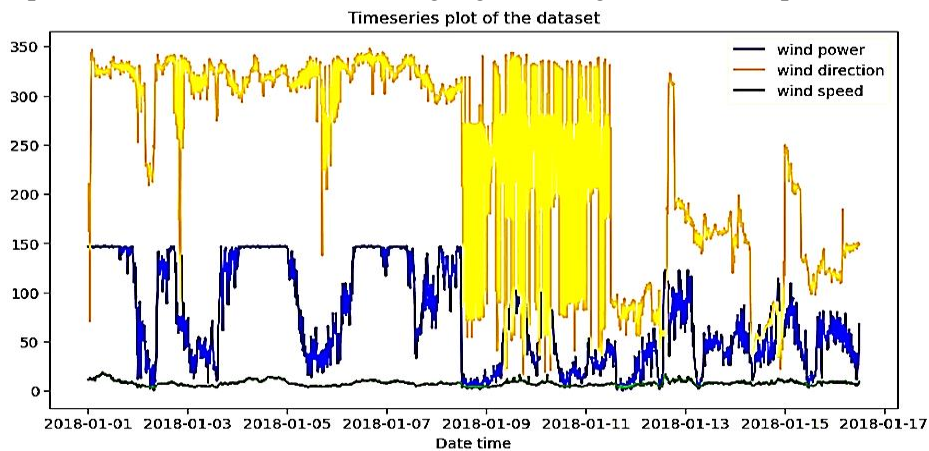

**Figure 1. Timeseries plot of the dataset**

Figure 1 displays the variability of each variable individually. However, given the significant values of wind power and wind direction, there is a need for an improved visualization that captures the evolution of wind speed and wind power over time. Figure 2, which excludes wind speed, offers a better representation of the dynamics between these variables [19]. It effectively illustrates the existing relationship between wind power and wind direction, providing valuable insights into their behavior over the observed time period.

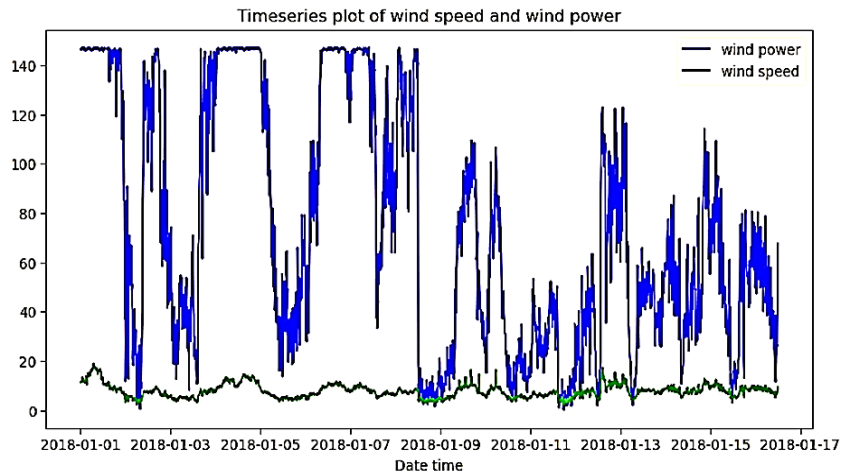

**Figure 2. Timeseries plot of wind speed and wind power**

Figures 1 and 2 enhance our understanding of the dataset by highlighting the absence of a clear trend, seasonality, and outliers. Moreover, the displayed stationarity of the variables indicates that additional pre-processing steps for time series analysis may not be necessary [20], [21]. The stationary nature of the variables suggests that they exhibit relatively consistent statistical properties over time, facilitating the application of time series modeling techniques without the need for extensive data transformations or adjustments.

**2.3. Algorithms**

The selection of algorithms for this study was based on their ability to handle both non-linearity and linearity in the dataset, as well as their compatibility for combination in bivariate multistep forecasting. Consequently, we considered the hybrid CNN-LSTM model [22], [23] and the Autoregressive model (or Linear Regression model) [24], as well as their combination, resulting in the CNN-LSTM-AR model. As there are numerous publications providing detailed information about these models, we will not delve into their specifics further. However, it is worth noting that combining different algorithms has the potential to improve forecasting accuracy. In the CNN-LSTM-AR model, the output of the CNN-LSTM model is utilized as the input for the Autoregressive Regression model to generate the final prediction. To enhance the performance of the proposed models, a feature engineering process was applied, focusing on the manipulation of the shaping of the input data. The details of this process are explained in the pseudo code provided in Pseudo Code 1.

*Pseudo code 1*

Assuming that:
CNN-LSTM = M1, Autoregressive model = M2, first shaping = S1, second shaping = S2, Y1 and Y2 = series to predict.

***For S1 applied on Y1 and Y2***:
    Prediction1 = M1[(Y1, Y2) [S1]]
    *S2 applied on Prediction1*

   Prediction 2 = M2[S2 applied on Prediction1]
**Metrics = metrics_name (test set of M1, prediction of Prediction 2)**

In the initial shaping process, considering the two series to predict, Y1 and Y2, for time series supervised learning, they are transformed into the following shapes: X = [n_samples, n_steps, n_features] and Y = [n_samples, n_features]. These shaped datasets are then divided into train, validation, and test sets. Subsequently, they are fed into the CNN-LSTM model for the first prediction, denoted as Prediction 1. The output of the CNN-LSTM model, denoted as M1, will have the shape [n_samples, n_features]. To prepare for the subsequent step, M1 is reshaped as follows: X is reshaped to [n_samples, n_steps, n_features], and Y is reshaped to [n_samples, n_features]. These reshaped datasets are then passed into the autoregressive model for the final prediction, referred to as M2. The evaluation metrics are computed based on the test set of M1 and the prediction of M2. These metrics provide insights into the accuracy and performance of the models in predicting the target variables.

## 2.4. Model building

### 2.4.1 Forecasting strategy

In the field of forecasting, there exist various strategies that can be employed [25]. For this study, the direct multistep strategy was investigated, wherein a separate model is built for each step in the prediction horizon [26], [27]. Two approaches were considered within this strategy. The first approach involved using an equal number of time steps for both models. In this case, the number of steps considered for the first model was the same as that for the second model. The second approach differed by increasing the number of steps for the first model, while keeping the number of steps for the second model at one. This allowed for a comparison between the performance of the two approaches and their respective modeling strategies.

The first approach can be explained as follows:

Final Prediction$_{(t+1)}$ = Forecasting$_{(t+1)}$ + **model$_1$** [*Pred obs$_{(t-1)}$, Pred obs$_{(t-2)}$, …, Pred obs$_{(t-n)}$*]
Final Prediction$_{(t+2)}$ = Forecasting$_{(t+2)}$ + **model$_2$** [*Pred obs$_{(t-2)}$, Pred obs$_{(t-3)}$, …, Pred obs$_{(t-n)}$*]
Final Prediction$_{(t+n)}$ = Forecasting$_{(t+n)}$ + **model$_n$** [*Pred obs$_{(t-n1)}$, Pred obs$_{(t-n2)}$, …, Pred obs$_{(t-nn)}$*]
   Where:
Final Forecasting $_{(t+1,…,t+n)}$ is the prediction of the proposed hybrid model at different time steps
Forecasting $_{(t+1…,t-n)}$ is the prediction at different timesteps for the CNN_LSTM model,
Model$_{1,…,n}$ is the Autoregressive model built using forecasting data as input with respective time step, and,
*Pred obs$_{(t+1,…t-nn)}$* is the observation of the forecasting of the CNN_LSTM.

The second approach is:
Final Prediction$_{(t+1)}$ = Forecasting$_{(t+1)}$ + **model$_1$** [*Pred obs$_{(t-1)}$, Pred obs$_{(t-2)}$, …, Pred obs$_{(t-n)}$*]
Final Prediction$_{(t+2)}$ = Forecasting$_{(t+2)}$ + **model$_1$** [*Pred obs$_{(t-1)}$, Pred obs$_{(t-2)}$, …, Pred obs$_{(t-n)}$*]
Final Prediction$_{(t+n)}$ = Forecasting$_{(t+n)}$ + **model$_1$** [*Pred obs$_{(t-1)}$, Pred obs$_{(t-2)}$, …, Pred obs$_{(t-n)}$*]
Where:
   Final Prediction $_{(t+1,…,t+n)}$ is the prediction of the proposed hybrid model at different time steps
Forecasting $_{(t+1…,t-n)}$ is the prediction at different timesteps for the CNN_LSTM model,
Model$_1$   is the Autoregressive model built using the forecasting data as input in keeping one step, and,

*Pred obs*$_{(t+1,...t-nn)}$ is the observation of the forecasting of the CNN_LSTM.

### 2.4.2. Model's structure

All the models were implemented using the Google Colab platform and Python 3.10. In addition, the CNN_LSTM was built using the Keras library and sklearn for the Autoregressive model. Table 1 provides the structure of each model.

**Table 1: Model's structure**

| Model | Layer type | Output shape | Loss | Activation function |
|---|---|---|---|---|
| CNN_LSTM [1] | Timedistributed [Conv1D] Timedistributed [Max pooling 1D] Timedistributed [Flatten] LSTM Dense Dense | (none, 1,1,350) (none, 1,1,350) (none, 1,350) (none, 350) (none, 300) (none, 2) | Mean Squared error | Relu |
| Autoregressive model [2] | **Parameters:** fit_intercept=**True,** normalize=**'deprecated',** copy_X=**True,** n_jobs=**None,** positive=**False** | | | |
| Proposed model | Combination of the structure of [1] and [2] | | | |

The mean squared error (MSE) was employed as the loss function and the 'relu' as activation function. To prevent overfitting or underfitting, several adjustments were made, including modifications in the number of epochs and the sizes of the training and validation sets. These adjustments were implemented based on the performance of the models and aimed to optimize their training process. The initial model structure, as presented in Table 3, was maintained across the different steps of the forecasting process. This consistency allowed for the evaluation of the models' performance over time. Furthermore, the performance of the CNN-LSTM model and the linear model were assessed separately. This evaluation provided a basis for comparison and allowed for a better understanding of the improvement achieved by the proposed model compared to these individual models.

### 2.5. Metrics

To strongly penalize error, the root mean squared error (RMSE) [28] was considered, and the precision of the models was evaluated using the r-squared[28]. Their respective equations are provided in 3 and 4.

$$RMSE = \sqrt{\frac{1}{m}\sum_{i=1}^{m}(X_i - Y_i)^2} \quad [3]$$

$$R^2 = \frac{\sum_{i=1}^{m}(X_i - Y_i)^2}{\sum_{i=1}^{m}(\bar{Y}_i - Y_i)^2} \; or \; \frac{Mean\,Squared\,Error}{Mean\,total\,Sum\,of\,Square} \quad [4]$$

### 2.6. Research design

The analysis conducted in this study was divided into two main sections. In the first phase, the prediction of the target variables was carried out using the CNN-LSTM and Autoregressive models independently. Each model was evaluated based on their respective metrics and performance. In the second phase, the focus shifted to the proposed model, which combined the CNN-LSTM and Autoregressive models. The performance of this combined model was assessed and compared to the individual models used in the first phase. The metrics obtained from this evaluation were analyzed and discussed to identify the factors contributing to their respective performances. Figure 3 provides a visual representation of the entire process, illustrating the progression from the independent prediction of the target variables to the evaluation and comparison of the models' metrics. This systematic approach allows for a comprehensive understanding of the performance and effectiveness of the proposed model in relation to the individual models.

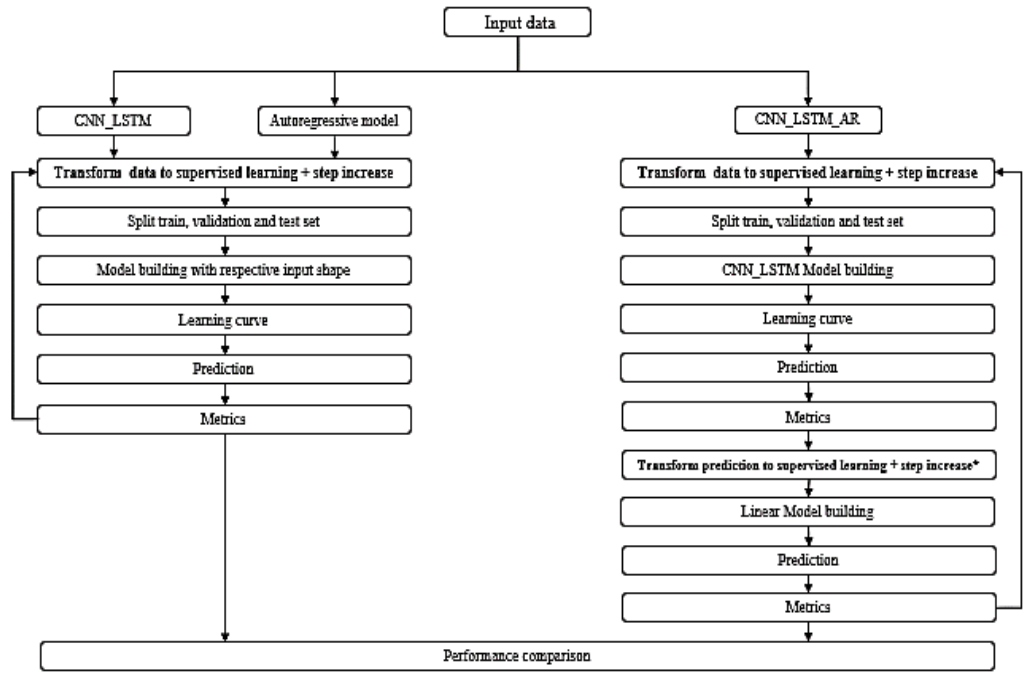

**Figure 3. Activity diagram**

*\* Increase steps only in the first shaping approach and in the first model of the second shaping approach*

In both stages of the analysis, the input data underwent a pre-processing step to ensure its suitability for supervised learning in time series forecasting. The data was divided into training and testing sets, and then fed into the models for prediction. To enhance the models' performance, the data was scaled using a MinMaxScaler and subsequently restored to its original form for the evaluation of metrics. The aforementioned procedures were repeated for each step and phase of the analysis, continuously monitoring the models' performance to detect any degradation such as overfitting or underfitting. Learning curves were utilized to assess the performance degradation and determine the optimal stopping point. Metrics were recorded and compared for each step of the experiment, up to the 384th step, which marked the conclusion of the analysis. This systematic approach ensured the evaluation and comparison of the models' performance across multiple steps, providing valuable insights into their predictive capabilities.

## 3. RESULT AND DISCUSSION

In the initial step, both the CNN-LSTM and Autoregressive models were designed to achieve high accuracy in predicting the target variables. Subsequently, additional steps were incrementally added while maintaining the original structure of each model. The performance of the models was closely monitored at each step, and any degradation in their performance was noted. Table 2 provides a comprehensive overview of the step-by-step performance of each model.

**Table 2: First approach: Model's performance by steps**

|       | Root mean squared error |          |             | Accuracy (%) |          |             |
|-------|-------------------------|----------|-------------|--------------|----------|-------------|
| Steps | CNN_LSTM                | AR Model | CNN_LSTM_AR | CNN_LSTM     | AR Model | CNN_LSTM_AR |
| 1     | 7.82                    | 8.68     | **7.67**    | 83.12        | 80.67    | **85.09**   |
| 3     | 7.67                    | 8.45     | **5.49**    | 85.13        | 81.45    | **91.24**   |
| 6     | 7.87                    | 8.48     | 11.66       | 84.04        | 81.21    | 66.22       |
| 12    | 7.72                    | 8.5      | 16.63       | 84.84        | 80.98    | 25.88       |
| 24    | 8.26                    | 8.57     | --          | 82.63        | 80.4     | --          |
| 48    | 9.07                    | 8.73     | --          | 79.39        | 79.15    | --          |
| 96    | 9.15                    | 8.99     | --          | 76.85        | 77.85    | --          |
| 192   | 10.67                   | 9.5      | --          | 64.6         | 75.16    | --          |
| 384   | 13.81                   | 12.19    | --          | 33.23        | 53.67    | --          |

In the first approach, where both models have similar shaping, the proposed model demonstrates superior performance compared to the other models in the short term (1st to 3rd steps). Specifically, the proposed model achieves accuracy rates of 85.09% and 91.24%, surpassing the CNN-LSTM model's accuracy rates of 83.12% and 85.13%, as well as the linear model's accuracy rates of 80.67% and 81.45%. On the other hand, the CNN-LSTM model performs better in the mid-term (6th to 24th steps), while the autoregressive model excels in the long-term (48th to 384th steps). These findings highlight the strengths and weaknesses of each model across different forecasting horizons. For a more comprehensive and visual representation of the results presented in Table 2, Figures 4 and 5 provide improved visualization, enabling a clearer understanding of the performance comparison among the models at various steps.

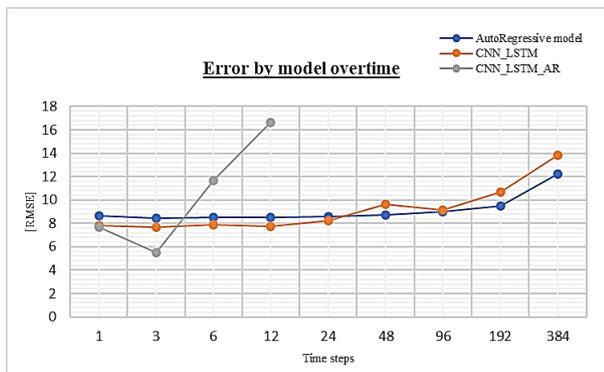  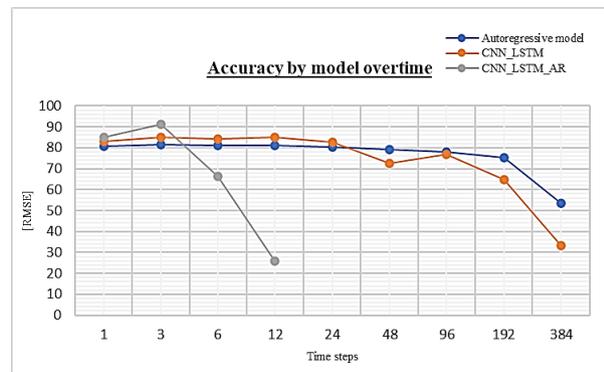

**Figure 4. Model's rmse over time**          **Figure 5. Model's accuracy over time**

Figures 4 and 5 provide valuable insights into the relationship between error and accuracy for each model. Figure 4 illustrates that an increase in error corresponds to a decrease in accuracy across the different forecasting steps. This observation underscores the challenge faced by the proposed approach in handling errors, particularly in the mid and long-term, resulting in its relatively poor performance compared to the other models. However, when the second approach is employed, involving different shaping for the two models within the proposed hybrid, the performance of the models is as follows in Table 3:

**Table 3. Second approach: Model's performance by steps**

| Steps | Root mean squared error | | | Accuracy (%) | | |
|---|---|---|---|---|---|---|
| | CNN_LSTM | AR Model | CNN_LSTM_AR | CNN_LSTM | AR Model | CNN_LSTM_AR |
| 1 | 7.82 | 8.68 | 7.64 | 83.12 | 80.67 | **83.36** |
| 3 | 7.67 | 8.45 | 7.92 | **85.13** | 81.45 | 84.71 |
| 6 | 7.87 | 8.48 | 7.71 | 84.04 | 81.21 | **84.61** |
| 12 | 7.72 | 8.5 | 7.77 | **84.84** | 80.98 | 83.92 |
| 24 | 8.26 | 8.57 | 7.9 | 82.63 | 80.4 | **83.91** |
| 48 | 9.07 | 8.73 | 10.93 | **79.39** | 79.15 | 73.71 |
| 96 | 9.15 | 8.99 | 10.31 | 76.85 | **77.85** | 73.65 |
| 192 | 10.67 | 9.5 | 12.76 | 64.6 | **75.16** | 54.96 |
| 384 | 13.81 | 12.19 | **15.5** | 33.23 | **53.67** | 42.61 |

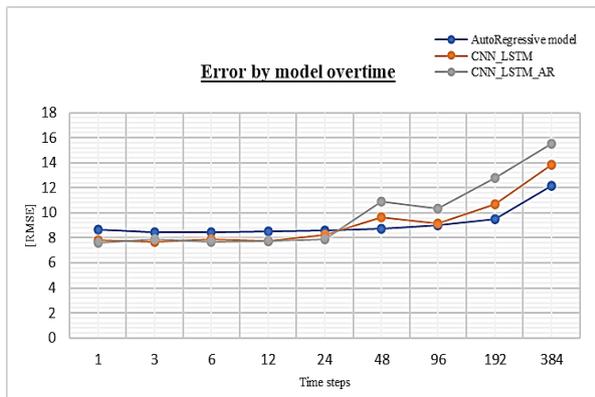

**Figure 6. Model's error time**

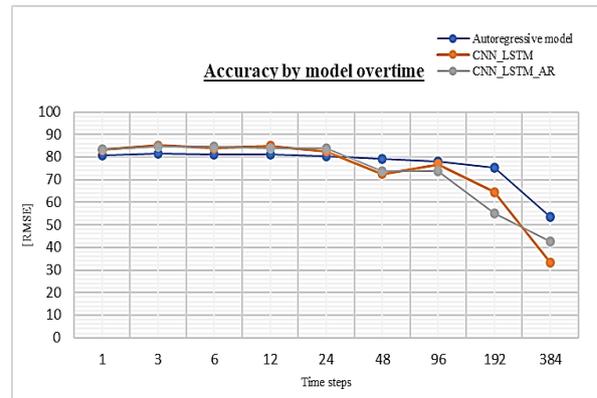

**Figure 7. Model's accuracy over time**

The second approach, which involves different shaping for the two models, demonstrates significant improvements compared to the first approach. While the first approach was limited to the short term and exhibited overfitting by the 12th step (as indicated in Table 2), the second approach extended the model's performance up to the 384th step (as shown in Table 3 and Figures 6 and 7). This indicates that the approach considering different shaping is more effective in mitigating errors and improving accuracy compared to the approach with similar shaping, which is only suitable for very short or short-term predictions. Understanding the relationship between error and accuracy is crucial in further analyzing the results and gaining deeper insights into the performance of the models.

### 3.1. Error analysis

#### 3.1.1. Error calculation

During the training process, both the CNN-LSTM and Autoregressive models utilize different approaches to calculate the error. The CNN-LSTM model employs backpropagation for the CNN and "backpropagation through time" for the LSTM. This involves adjusting the weights based on the partial derivatives of the error with respect to each weight, optimizing the model through gradient descent [29]. On the other hand, training a linear model involves finding the best set of parameters (coefficients) that minimize the difference between the predicted output and the actual output [30]. These distinct error calculation methods yield different results, even when applied to the same dataset. In cases where there is a limited amount of data, the error calculation performed by the linear model appears to be better than that of the CNN-LSTM model, which requires a larger dataset to better capture trends. The proposed model takes advantage of both error calculation approaches to enhance its performance and address the limitations of individual models.

### 3.1.2. Error accumulation

When forecasting for a longer number of time steps, the accumulation of errors becomes a significant challenge in time series analysis [24], [29]. This phenomenon, known as error accumulation, occurs as the biases and variances from previous steps propagate to subsequent steps. The effectiveness of the proposed model in minimizing error relies on the individual models' ability to handle such errors. In the absence of noise, the proposed model demonstrates superior short-term performance compared to the individual models evaluated separately. However, as the number of time steps increases, the presence of noise hampers the efficacy of both backpropagation and residual computation in minimizing error, as they are susceptible to noise interference [31], [32]. The CNN-LSTM model excels in capturing high-level features and temporal dependencies within the input data [33]. In contrast, the linear model focuses on capturing low-level features and linear relationships within the data. By combining these models, it is possible to capture the complex and nonlinear relationships between the input features and the target variable more effectively. When applying similar shaping in both models, the proposed model achieves excellent performance in the short term due to the limited presence of noise. However, this shaping approach may remove important information or introduce artificial features in the data, leading to suboptimal performance. Conversely, employing different shaping approaches allows the CNN-LSTM model to filter out irrelevant information and focus on the most informative features, thereby reducing noise and variability in the input data and making it more suitable for the linear model. Moreover, reshaping the output of the CNN-LSTM model helps mitigate overfitting and enhances the model's generalization performance. By simplifying the data representation through reshaping, the risk of overfitting is reduced, and the model becomes more capable of generalizing to new, unseen data. Therefore, selecting an appropriate shaping approach is crucial to ensure reliable predictions. Figures 8 and 9, provide a comprehensive comparison and visualization of the two shaping approaches, highlighting the significance of the proposed shaping method.

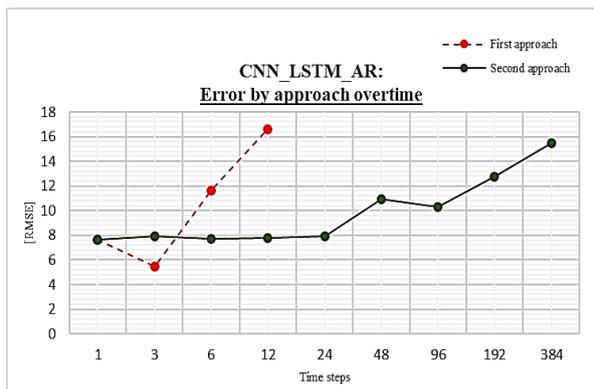
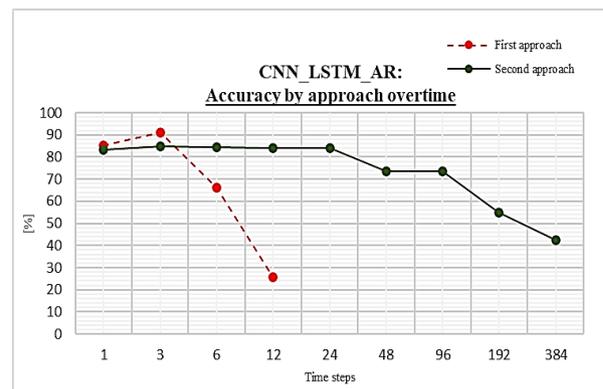

**Figure 8. RMSE by approach overtime**         **Figure 9. Accuracy by approach overtime**

These figures present the temporal progression of error and accuracy for the proposed hybrid model, taking into account the shaping approach. The incorporation of a well-designed feature combination in the suggested model leads to improved prediction accuracy for both short-term and long-term forecasts. This outcome indicates the potential for further exploration and research in this field, as it demonstrates the effectiveness of the proposed model and its ability to generate more accurate predictions.

## 4. CONCLUSION

The selection of an appropriate forecasting algorithm for a specific time series can be challenging, as it depends on both the characteristics of the time series itself and the capabilities of the forecasting algorithms. This study addresses this challenge by employing a feature engineering process that focuses on differing the shape of the data in each model. Basically, steps were increased before passing data in the first model, then, before passing to the second model, it was returned to the one step ahead shaping. Specifically, the performance of a hybrid model combining the CNN LSTM and Autoregressive model is analyzed for bivariate multistep forecasting, and compared to the performance of each individual model. The results provide insights into their relative performances and shed light on the strengths and weaknesses of the proposed approach. The suggested method demonstrates superior short-term performance compared to existing models when using similar shaping techniques, while also showing the potential to improve mid- and long-term performance by employing different shaping techniques. This approach proves to be effective in mitigating the accumulation of errors over time, making it suitable for bivariate forecasting tasks such as the one considered in this study. It offers a promising avenue for enhancing forecasting accuracy in the near term and minimizing the impact of error accumulation. It is important to note that the outcomes of this research are limited by the constraints imposed at the beginning of the study, which were intended to facilitate a better understanding of the capabilities of the models within the same context. The constraints serve as a means to assess their respective strengths and weaknesses and provide a clear interpretation of the performance of the proposed hybrid model. This research primarily focuses on comprehending the behavior of the selected models in the context of the feature engineering task conducted. Further research could delve deeper into the understanding of this approach by exploring various combinations (different input shaping) and offering explanations for the observed results.